\documentclass[10pt,twocolumn,letterpaper]{article}

\usepackage{cvpr}
\usepackage{times}
\usepackage{epsfig}
\usepackage{graphicx}
\usepackage{amsmath}
\usepackage{amssymb}
\usepackage{comment}
\usepackage{float}
\usepackage{tabularx, booktabs}
\usepackage{multirow}
\usepackage{epstopdf}
\usepackage{authblk}
\usepackage{lipsum}

\newcommand\blfootnote[1]{%
  \begingroup
  \renewcommand\thefootnote{}\footnote{#1}%
  \addtocounter{footnote}{-1}%
  \endgroup
}

\newcolumntype{x}[1]{>{\centering\let\newline\\\arraybackslash\hspace{0pt}}p{#1}}

\usepackage[pagebackref=true,breaklinks=true,colorlinks,bookmarks=false]{hyperref}

\cvprfinalcopy 

\def\cvprPaperID{2996}

\ifcvprfinal\fi
\begin{document}

\title{Blind Geometric Distortion Correction on Images Through Deep Learning}



\author{Xiaoyu Li$^{1}$ \qquad Bo Zhang$^{1}$ \qquad Pedro V. Sander$^{1}$ \qquad Jing Liao$^{2}$ \vspace{1pt}\\
$^{1}$The Hong Kong University of Science and Technology \qquad $^{2}$City University of Hong Kong \qquad\qquad\\
}

\maketitle
\thispagestyle{empty}

\begin{abstract}
We propose the first general framework to automatically correct different types of geometric distortion in a single input image. Our proposed method employs convolutional neural networks (CNNs) trained by using a large synthetic distortion dataset to predict the displacement field between distorted images and corrected images. A model fitting method uses the CNN output to estimate the distortion parameters, achieving a more accurate prediction. The final corrected image is generated based on the predicted flow using an efficient, high-quality resampling method. Experimental results demonstrate that our algorithm outperforms traditional correction methods, and allows for interesting applications such as distortion transfer, distortion exaggeration, and co-occurring distortion correction.
\end{abstract}

\blfootnote{\href{https://xiaoyu258.github.io/projects/geoproj}{Webpage: https://xiaoyu258.github.io/projects/geoproj} }

\section{Introduction}

\begin{figure}
	\centering
	\includegraphics[width=\linewidth]{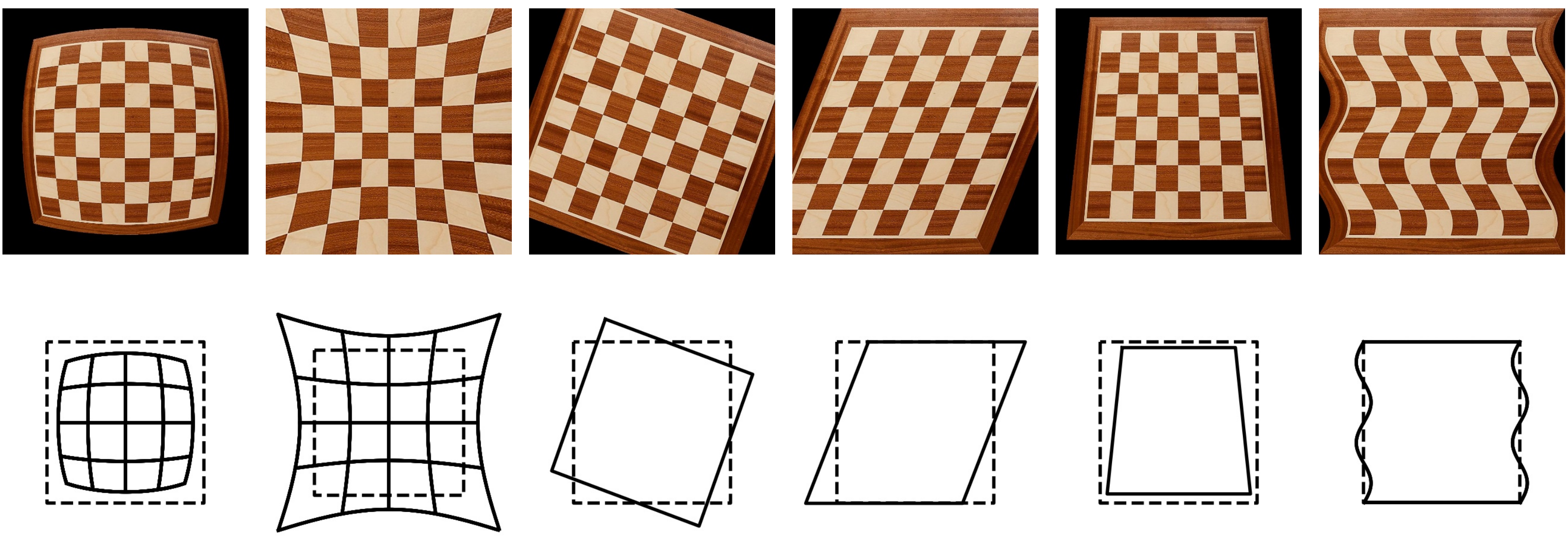}
	\begin{tabular}{x{0.1\linewidth}x{0.12\linewidth}x{0.12\linewidth}x{0.12\linewidth}x{0.12\linewidth}x{0.1\linewidth}}
		$\mathcal{B}$ & $\mathcal{P}i$ & $\mathcal{R}$ & $\mathcal{S}$ & $\mathcal{P}$ & $\mathcal{W}$\\
	\end{tabular}
	\caption{Our system is trained to correct barrel distortion ($\mathcal{B}$), pincushion ($\mathcal{P}i$), rotation ($\mathcal{R}$), shear ($\mathcal{S}$), perspective ($\mathcal{P}$) and wave distortion ($\mathcal{W}$).}
	\label{fig:distortion}
\end{figure}

Geometric distortion is a common problem in digital imagery and occurs in a wide range of applications. It can be caused by the acquisition system (e.g., optical lens, imaging sensor), imaging environment (e.g., motions of the platform or target, viewing geometry) and image processing operations (e.g., image warping). For example, camera lenses often suffer from optical aberrations, causing barrel distortion ($\mathcal{B}$), common in wide angle lenses, where the image magnification decreases with distance from the optical axis, and pincushion distortion ($\mathcal{P}i$), where it increases. While lens distortions are intrinsic to the camera, extrinsic geometric distortions like rotation ($\mathcal{R}$), shear ($\mathcal{S}$) and perspective distortion ($\mathcal{P}$) may also arise from the improper pose or the movement of cameras. Furthermore, a wide number of distortion effects, such as wave distortion ($\mathcal{W}$), can be generated by image processing tools. We aim to design an algorithm that can automatically correct images with these distortions and can be generalized to a wide range of distortions easily (see Figure~\ref{fig:distortion}). 

\begin{figure}
\begin{center}
    \includegraphics[width=\linewidth]{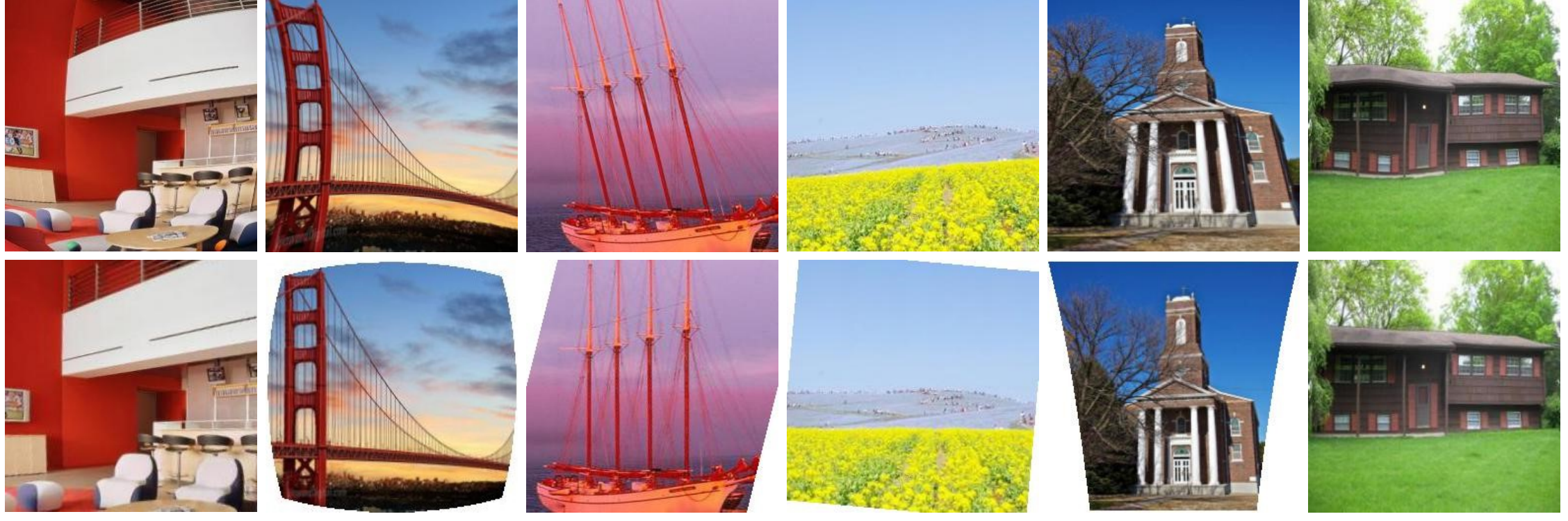}
\end{center}
    \vspace{-0.3cm}
	\caption{Our proposed learning-based method can blindly correct images with different types of geometric distortion (first row) providing high-quality results (second row).}
    \label{fig:teaserfigure}
\end{figure}

Geometric distortion correction is highly desired in both photography and computer vision applications. For example, lens distortion violates the pin-hole camera model assumption which many algorithms rely on. Second, remote sensing images usually contain geometric distortions that cannot be used with maps directly before correction~\cite{toutin2004geometric}. Third, skew detection and correction is an important pre-processing step in document analysis and has a direct effect on the reliability and efficiency of the segmentation and feature extraction stages~\cite{al2009skew}. Finally, photos often contain slanted buildings, walls, and horizon lines due to improper camera rotation. Our visual system expects man-made structures to be straight, and horizon lines to be horizontal~\cite{lee2012automatic}.

Completely blind geometric distortion correction is a challenging problem, which is under-constrained given that the input is only a single distorted image. Therefore, many correction methods have been proposed by using multiple images or additional information. Multiple views methods~\cite{barreto2005fundamental, hartley2007parameter, kukelova2011minimal} for radial lens distortion use point correspondences of two or more images. These methods can achieve impressive results. However, they cannot be applied when multiple images under camera motion are unavailable.

To address these limitations, distortion correction from a single image has also been explored. Methods for radial lens distortion based on the plumb line approach~\cite{wang2009simple, bukhari2013automatic, thormahlen2003robust} assume that straight lines are projected to circular arcs in the image plane caused by radial lens distortion. Therefore, accurate line detection is a very important aspect for the robustness and flexibility of these methods. Correction methods for other distortions~\cite{gallagher2005using,lee2012automatic,chaudhury2015automatic,santana2017automatic} also rely on the detection of special low-level features such as vanishing points, repeated textures, and co-planar circles. But these special low-level features are not always frequent enough for distortion estimation in some images, which greatly restrict the versatility of the methods. Moreover, all of the methods focus on a specific distortion. To our knowledge, there is no general framework which can address different types of geometric distortion from a single image. 

In this paper, we propose a learning-based method to achieve this goal. We use the displacement field between distorted images and corrected images to represent a wide range of distortions. The correction problem is then converted to the pixel-wise prediction of this displacement field, or flow, from a single image. Recently, CNNs have become a powerful method in many fields of computer vision and outperform many traditional methods, which motivated us to use a similar network structure for training. The predicted flow is then further improved by our model fitting methods which estimate the distortion parameters. Lastly, we use a modified resampling method to generate the output undistorted image from the predicted flow.

Overall, our learning-based method does not make strong assumptions on the input images while generating high-quality results with few visible artifacts as shown in Figure~\ref{fig:teaserfigure}. Our main contribution is to propose the first learning-based methods to correct a wide range of geometric distortions blindly. More specifically, we propose:

\begin{enumerate}
\item A single-model network, which implicitly learns the distortion parameters given the distortion type.
\item A multi-model network, which performs type classification jointly with flow regression without knowing the distortion type, followed by an optional model fitting method to further improve the accuracy of the estimation.
\item A new resampling method based on an iterative search with faster convergence. 
\item Extended applications that can directly use this framework, such as distortion transfer, distortion exaggeration, and co-occurring distortion correction.
\end{enumerate}

\section{Related Work}
\paragraph{Geometric distortion correction.} For camera lens distortions, pre-calibration techniques have been proposed for correction with known distortion parameters~\cite{tardif2009calibration,duane1971close,tsai1987versatile,heikkila1997four,zhang1999flexible}. However, they are unsuitable for zoom lenses, and the calibration process is usually tedious. On the other hand, auto-calibration methods do not require special calibration patterns and automatically extracts camera parameters from multi-view images~\cite{fitzgibbon2001simultaneous,hartley2007parameter,kukelova2011minimal,ramalingam2010generic,henrique2013radial}. But for many application scenarios, multiple images with different views are unavailable. To address these limitations, automatic distortion correction from a single image has gained more research interest recently. Fitzgibbon~\cite{fitzgibbon2001simultaneous} proposes a division model to approximate the radial distortion curve with higher accuracy and fewer parameters. Wang \etal~\cite{wang2009simple} studied the geometry property of straight lines under the division model and proposed to estimate the distortion parameters through arc fitting. Since plumb line methods rely on robust line detection, Aleman-Flores \etal~\cite{aleman2014automatic} used an improved Hough Transform to improve the robustness while Bukhari and Dailey~\cite{bukhari2013automatic} proposed a sampling method that robustly chooses the circular arcs and determines distortion parameters that are insensitive to outliers. 

For other distortions, such as rotation and perspective, most of the image correction methods rely on the detection of low-level features such as vanishing point, repeated textures, and co-planar circles~\cite{gallagher2005using,lee2012automatic, chaudhury2015automatic, santana2017automatic}. Recently, Zhai \etal~\cite{zhai2016detecting} proposed to use deep convolutional neural networks to estimate the horizon line by aggregating the global image context with the clue of the vanishing point. Workman \etal~\cite{workman2016horizon} goes further and directly estimates the horizon line in the single image. Unlike these specialized methods, our approach is generalizable for multi-type distortion correction using a single network.

\paragraph{Deformation estimation.} There has been recent work on automatic detection of geometric deformation or variations in a single image. Dekel \etal~\cite{dekel2015revealing} use a non-local variations algorithm to automatically detect and correct small deformations between repeating structures from a single image. Wadhwa \etal~\cite{wadhwa2015deviation} fit parametric models to compute the geometric deviations and exaggerate the departure from ideal geometries. Estimating deformations has also been studied in the context of texture images \cite{kim2012symmetry, hays2006discovering, park2009deformed}. None of these techniques are learning-based and are mostly for specialized domains.


\paragraph{Neural networks for pixel-wise prediction.} Recently, convolutional neural networks have been used in many pixel-wise prediction tasks from a single image, such as semantic segmentation~\cite{long2015fully}, depth estimation~\cite{eigen2014depth} and motion prediction~\cite{walker2015dense}. One of the main problems for dense prediction is how to combine multi-scale contextual reasoning with the full-resolution output. Long \etal~\cite{long2015fully} proposed fully convolutional networks which popularized CNNs for dense predictions without fully connected layers. Some methods focus on dilated or atrous convolution~\cite{yu2015multi, chen2018deeplab} which supports exponential expansion the receptive field and systematically aggregate multi-scale contextual information without losing resolution. Another strategy is to use the encoder-decoder architecture~\cite{noh2015learning, badrinarayanan2015segnet, ronneberger2015u}. The encoder gradually reduces the spatial dimension to increase the receptive field of the neuron, while the decoder maps the low-resolution feature maps to full input resolution maps. Noh \etal~\cite{noh2015learning} developed deconvolution and unpooling layers for the decoder part. Badrinarayanan \etal~\cite{badrinarayanan2015segnet} used pooling indices to connect the encoder and the corresponding decoder, making the architecture more memory efficient. Another popular network is U-net~\cite{ronneberger2015u}, which uses the skip connections to combine the contracting paths with the upsampled feature maps. Our networks use an encoder-decoder architecture with residual connection design and achieve more accurate results.

\section{Network Architectures}
\label{sec:net}

\begin{figure*}
    \centering
    \includegraphics[width=\linewidth]{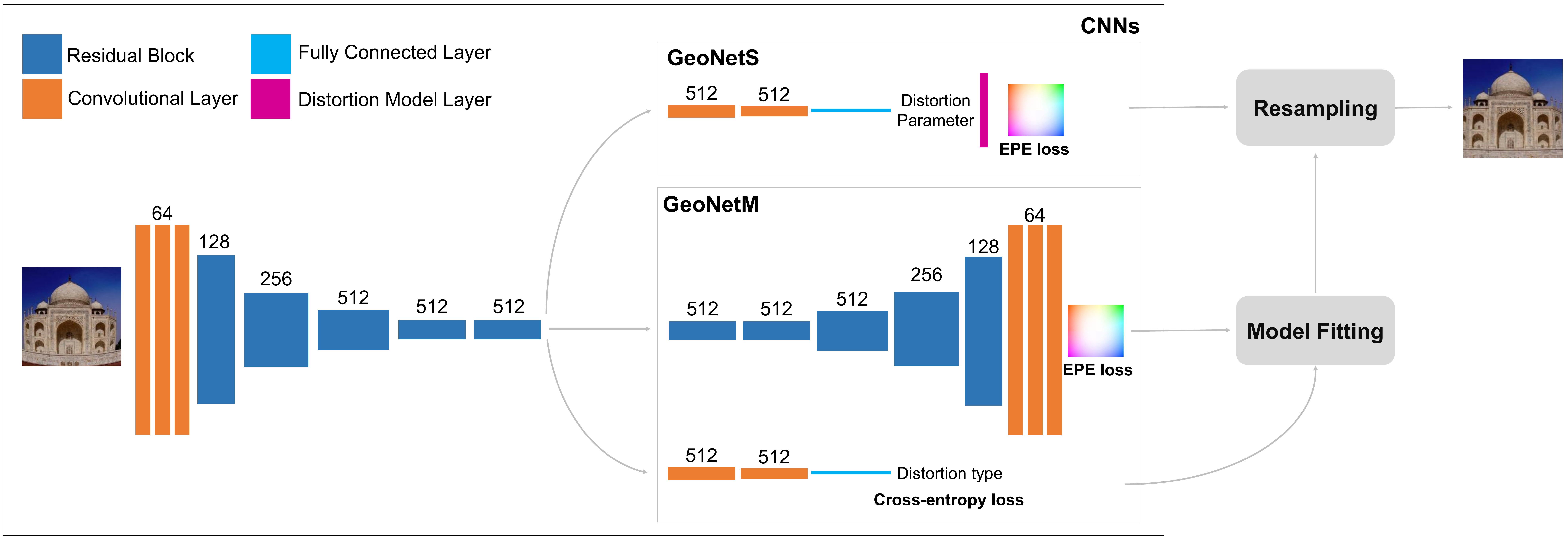}
    \caption{Overview of our entire framework, including the single-model (GeoNetS) and multi-model (GeoNetM) distortion networks (Section~\ref{sec:net}), and resampling (Section~\ref{sec:resample}). Each box represents some conv layers, with vertical dimension indicating feature map spatial resolution, and horizontal dimension indicating the output channels of each conv layer in the box.}
    \label{fig:architecture}
\end{figure*}

Geometrically distorted images usually exhibit unnatural structures that can serve as clues for distortion correction. As a result, we presume that the network can potentially recognize the geometric distortions by extracting the features from the input image. We, therefore, propose a network to learn the mapping from the image domain $\mathcal{I}$ to the flow domain $\mathcal{F}$. The flow is the 2D vector field that specifies where pixels in the input image should move in order to get the corrected image. It defines a non-parametric transformation, thus being able to represent a wide range of distortions. Since the flow is a forward map from the distorted image to the corrected image, a resampling method is needed to produce the final result.

This strategy follows learning methods of other applications which have observed that it is often simpler to predict the transformation from input to output rather than predicting the output directly (e.g., ~\cite{gharbi2015transform, isola2017image}). Thus, we designed our architecture to learn an intermediate flow representation. Additionally, the forward mapping indicates where each pixel with a known color in the distorted image maps to. Therefore, all pixels in the input image learn a distortion flow prediction directly associated to them, which would not be the case if we were attempting to learn a backward mapping, where some input regions could not have correspondences. It can be a serious problem when the distortion changes the image shape greatly. Furthermore, our resampling method that is required to generate the final image is fast and accurate. 

We propose two networks by considering whether the user has prior knowledge of the distortion type. Our networks are trained in a supervised manner. Therefore, we first introduce how the paired datasets have been constructed (Section~\ref{sec:data}) and then we introduce our two networks, for single-model and multi-model distortion estimation (Sections~\ref{sec:net1} and~\ref{sec:net2}, respectively).

\subsection{Dataset construction}
\label{sec:data}
We generate the distorted image flow pair by warping an image with a given mapping, thus constructing the distorted image dataset $\mathcal{I}$ and its corresponding distortion flow dataset $\mathcal{F}$, where $I_j\in\mathcal{I}$ and $F_j\in\mathcal{F}$ are paired. 

We consider six geometric distortion models in our network. However, the architecture is not specialized to these types of distortion and can potentially be further extended. Each distortion type $\beta=1,...,6$ has a model $\mathcal{M}^\beta$, which defines the mapping from the distorted image lattice to the original one. Bilinear interpolation is used if the corresponding point in the original image is not on the integer grid. The flow $F = \mathcal{M}^\beta( \rho^\beta), F \in \mathcal{F}$, is generated in the meantime to record how the pixel in the distorted image should be moved to the corresponding point in the original image. $\rho^\beta$ is the distortion parameter that controls the distortion effect. For instance, in the rotation distortion model, $\rho^\beta$ is the rotation angle while in the barrel and pincushion model, $\rho^\beta$ represents the parameter in Fitzgibbon's single parameter division model~\cite{fitzgibbon2001simultaneous}. All distortion parameters $\rho^\beta$ in different distortion models are randomly sampled using a uniform distribution within a specified range. As Figure~\ref{fig:distortion} shows, the geometric distortions change the image shapes. Thus we crop the images and flows to remove empty regions.

\subsection{Single-model distortion estimation}
\label{sec:net1}
We first introduce a network $\mathcal{N}^\beta$ parameterized by $\theta^\beta$ to estimate the flow for distorted images with a known distortion type $\beta$. $\mathcal{N}^\beta$ learns the mapping from $\mathcal{I}^\beta$ to $\mathcal{F}^\beta$ with sub-datasets where $\mathcal{I}^\beta \subset \mathcal{I}$ and $\mathcal{F}^\beta \subset \mathcal{F}$ are sub-domains containing the images and flows of distortion type $\beta$.

\paragraph{Architecture.} A possible architecture choice is to directly regress the distortion flow according to the ground truth with an auto-encoder-like structure. However, the network would only be optimized with the pixel-wise flow error, without taking advantage of global constraints imposed by the known distortion model. Instead, we design a network to first predict the model parameter $\rho^\beta$ directly. This parameter is then used to generate the flow $F = \mathcal{M}^\beta( \rho^\beta)$ in the network. Though the network should implicitly learn the distortion parameter, there is no explicit constraint for the network to do so exactly.

The network architecture, referred to as GeoNetS, is shown in Figure~\ref{fig:architecture}. It has three conv layers at the very beginning and five residual blocks (\cite{he2016deep}) to gradually downsize the input image and extract the features. Each residual contains two conv layers and has a shortcut connection from input to output. The shortcut connection helps ease the gradient flow, achieving a lower loss according to our experiments. Downsampling in spatial resolution is achieved using conv layers with a stride of 2 and $3 \times 3$ kernels. Batch normalization layers and ReLU function are added after each conv layer, which significantly improves training.

After the residual blocks, two conv layers are used to downsize the features further, and a fully-connected layer converts the 3D feature map into a 1D vector $\rho^\beta$. With the distortion parameter $\rho^\beta$, the corresponding distortion model $\mathcal{M}^\beta$ analytically generates the distortion flow. The network is optimized with the pixel-wise flow error between the generated flow and the ground truth. 

\paragraph{Loss} We train the network to minimize the loss $\mathcal{L}$, which measures the distance between the estimated distortion flow and the ground truth flow:
\begin{equation}
\begin{aligned}
&^*\theta^\beta = \arg\min_{\theta^\beta}\mathcal{L}(\mathcal{N}^\beta(I;\theta^\beta), F) \\
&\mathcal{N}^\beta(I;\theta^\beta) = \mathcal{M}^\beta(n^\beta(I;\theta^\beta))
\end{aligned}
\end{equation}
where $n^\beta$ is the sub-network of $\mathcal{N}^\beta$, represents the part to regress the distortion parameter implicitly. Here we choose the endpoint error (EPE) as our loss function. The EPE is defined as the Euclidean distance between the predicted flow vector and the ground truth averaged over all pixels. Because the estimated distortion flow is explicitly constrained by the distortion model, it is naturally smooth. 

Since the geometric distortion models $\mathcal{M}^\beta$ we consider are differentiable, the backward gradient of each layer can be computed using the chain rule:
\begin{equation}
\frac{\partial \mathcal{L}}{\partial \theta^\beta} = \frac{\partial \mathcal{L}}{\partial \mathcal{M}^\beta } \frac{\partial \mathcal{M}^\beta} {\partial n^\beta}\frac{\partial n^\beta}{\partial \theta^\beta}
\end{equation}
Our trained network can estimate the distortion flow blindly from an input image for each distortion type and achieve comparable performance as traditional methods. 

\subsection{Multi-model distortion estimation}
\label{sec:net2}
The GeoNetS network is only able to capture a specific distortion type with a distortion model at a time. For a new type, the entire network has to be retrained. Furthermore, the distortion type and model can be unknown in some cases. In view of these limitations, we designed a second network for multi-model distortion estimation. However, since the distortion model and the parameters $\rho^\beta$ can vary drastically across types, it is impossible to train a multi-model network with the model constraints. We train a network to regress the distortion flow without model constraints and at the same time classify the distortion type. The network is illustrated in Figure~\ref{fig:architecture}. The multi-model network $\mathcal{N}$ parameterized by $\theta$ is jointly trained for two tasks. The first task estimates the distortion flow, learning the mapping from the image domain $\mathcal{I}$ to the flow domain $\mathcal{F}$. The second task classifies the distortion type, learning the mapping from image domain $\mathcal{I}$ to type domain $\mathcal{T}$. 

\paragraph{Architecture} The entire network adopts an encoder-decoder structure, which includes an encoder part, a decoder part, and a classification part. The input image is fed into an encoder to encode the geometric features and capture the unnatural structures. Then two branches follow: In the first branch, a decoder is used to regress the distortion flow, while in the second branch a classification subnet is used to classify the distortion type. The encoder part is the same as GeoNetS, and the decoder part is symmetric to the encoder. Downsampling/Upsampling in spatial resolution is achieved using conv/upconv layers with a stride of 2. The classification part also has two conv layers to downsize the features further, and a fully-connected layer converts the 3D feature map into a 1D score vector of each type.

\paragraph{Loss} We use the EPE loss $\mathcal{L}_{\text{flow}}$ in the flow regression branch, and a cross entropy loss in the classification branch. The two branches are jointly optimized by minimizing the total loss:
\begin{equation}
^*\theta = \arg \min_{\theta} (\mathcal{L}_{\text{flow}}+ \lambda\mathcal{L}_{\text{class}})
\label{eq:l3}
\end{equation}
where the weight $\lambda$ provides a trade-off between the flow prediction and the distortion type classification.

We observe that jointly learning the distortion type helps reduce the flow prediction error as well. These two branches share the same encoder, and the classification branch helps the encoder learn the geometric features for different distortion types better. Please refer to Section~\ref{sec:results} for direct comparisons.

\paragraph{Model fitting}
Our multi-model network simultaneously predicts the flow and the distortion type from the input image. Based on this information, we can estimate the actual distortion parameters in the model and regenerate the flow to obtain a more accurate result. 

The Hough Transform is a widely used technique to extract features in an image. It is robust to noise by eliminating the outliers in the flow using a voting procedure. Moreover, it is a non-iterative approach. Each data point is treated independently, and therefore parallel processing of all points is possible. This makes it more computationally efficient. 

For an input image $I$, given its distortion type $\beta$ and distortion flow $\mathcal{N}(I;^*\theta)$ predicted by our network, we want to fit the corresponding distortion model $\mathcal{M}^\beta$ with the distortion parameter $\rho^\beta$. In our scenario, we map each data point $\mathcal{N}_{i j}$ in flow $\mathcal{N}(I;^*\theta)$ at position $(i, j)$ to a point in the distortion parameter space. The transform is given by
\begin{equation}
\rho_{ij} = \mathcal{M}^{-1}(\mathcal{N}_{i j})
\end{equation}

We assume the distortion parameter $\rho$ has a range from $\rho_{min}$ to $\rho_{max}$ and split the range into $M$ cells uniformly. All the points $\rho_{ij}$ belong to a cell according to the parameter values. The cell receiving the maximum number of counts determine the best fitting result, and the final result is the average of all the points in this cell. We let $M = 100$ in our experiments.

Once model fitting is completed, we have a refined and smoother flow $F=\mathcal{M}^{\beta}(\rho^\beta)$. With model fitting, the efficiency in correcting of higher resolution images can be greatly improved. This is because we can estimate the flow and obtain the distortion parameter $\rho$ at a much smaller resolution, and generate the full resolution flow directly according to the distortion parameter. 

\section{Resampling}
\label{sec:resample}

\begin{figure}
	\centering
    \footnotesize
	\includegraphics[width=\linewidth]{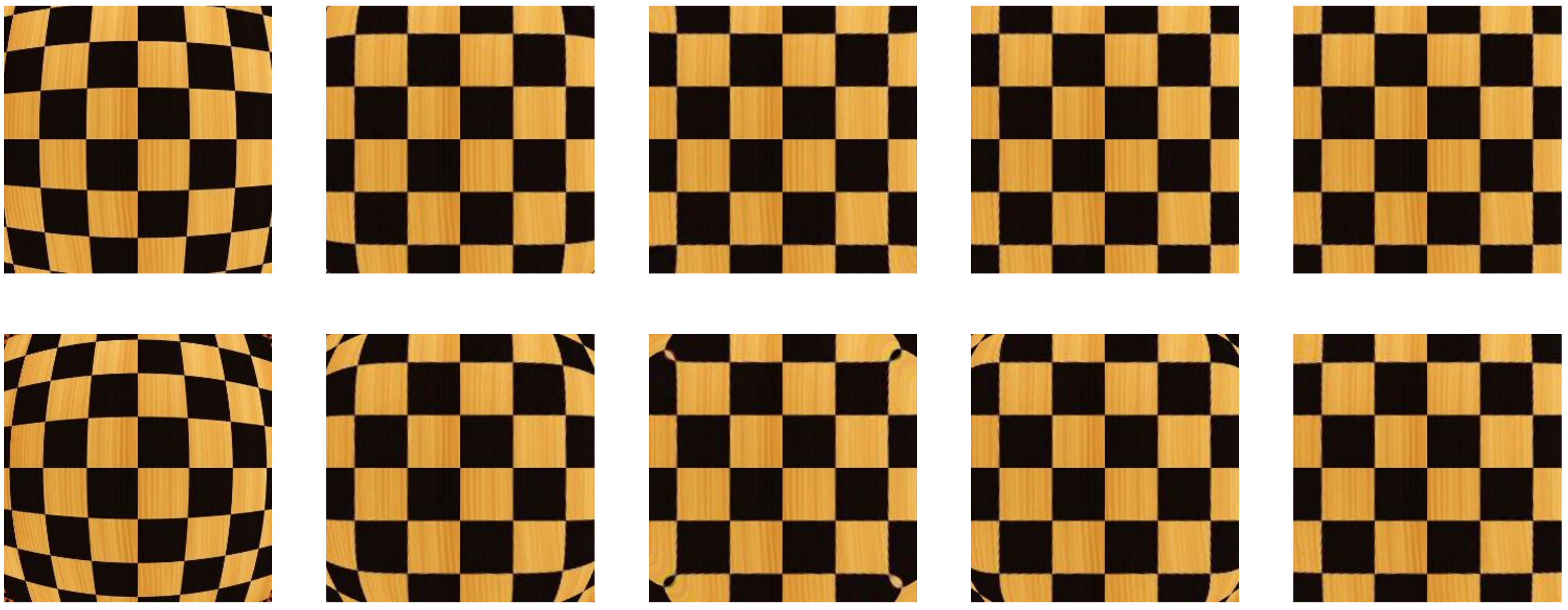}
     \begin{tabular}{x{0.14\linewidth}x{0.16\linewidth}x{0.16\linewidth}x{0.16\linewidth}x{0.14\linewidth}}
		Distorted & IS (5) & IS (10) & IS (15) & Ours (5) \\
	\end{tabular}   
	\caption{Comparison of the convergence using the traditional iterative search (IS) and our approach.  Two examples with different distortion levels are used. Our method converges to good resampling result with 5 iterations.}
	\label{fig:IS_images}
\end{figure}

Given the distortion flow, we employ a pixel evaluation algorithm to determine the backward-mapping and resample the final undistorted image. The approach is inspired by the bidirectional iterative search algorithm in~\cite{Yang2011Bidirectional}. Unlike mesh rasterization approaches, this iterative method runs entirely independently and in parallel for each pixel, fetching the color from the appropriate location of the source image.

The traditional backward mapping algorithm of~\cite{Yang2011Bidirectional} seeks to find a point $p$ in the source image that maps to $q$. Since we only have the forward distortion flow, this method essentially inverts this mapping using an iterative search until the location $p$ converges:

\begin{equation}
\begin{split}
p^{(0)} &= q \\
p^{(i+1)} &= q - f(p^{(i)})
\label{iter}
\end{split}
\end{equation}

\noindent where $f(p)$ is the computed forward flow from the source pixel $p$ to the undistorted image.

Since the application in this paper often involves large, smooth distortions, we propose a modification that significantly improves the convergence rate and quality. The traditional method initializes $p$ based on the flow at $q$. More specifically, $p^{(1)} = q - f(p^{(0)})$. If $f(p^{(1)}) \approx f(p^{(0)})$, then the iterative search converges quickly. However, in the presence of large distortions, $\| f(p^{(0)})\|$ is large, $p^{(0)}$ and $p^{(1)}$ are distant, and thus, $f(p^{(1)})$ and $f(p^{(0)})$ can be very different, making it a poor initialization and decreasing conversion speed.

Instead of assuming that the flow in $p^{(0)}$ and $p^{(1)}$ are the same, we compute the local derivative of the flow at $p^{(0)}$ using the finite difference method, and use this derivative to estimate the flow at $p^{(1)}$. We let $f_x(p)$ and $f_y(p)$ represent the horizontal and vertical flow respectively. Formally,

\begin{equation}
\frac{d f_x}{d x} = \frac{f_x(p^{(0)}_{nx}) - f_x(p^{(0)})}{(x + 1) - x}
\label{eq:derivative1}
\end{equation}

\noindent where $p^{(0)}$ is at coordinates $(x,y)$, and its horizontal pixel neighbor $p_{nx}$ is at coordinates $(x+1,y)$. We then use this derivative to approximate the flow at $p^{(1)} = (x', y')$:

\begin{equation}
\frac{d f_x}{d x} = \frac{f_x(p^{(1)}) - f_x(p^{(0)})}{x' - x}
\label{eq:derivative2}
\end{equation}

\noindent By the definition of forward flow, we have $f_x(p^{(1)}) = x - x'$. Therefore we can compute $x'$ combining Equation~\ref{eq:derivative1} and Equation~\ref{eq:derivative2}:

\begin{equation}
x' = x - \frac{f_x(p^{(0)})}{1 + f_x(p^{(0)}_{nx}) - f_x(p^{(0)})} \\
\end{equation}

We compute $y'$ similarly and proceed with the iterative search. Note that we only use this finite difference method in the first iteration to get a coarse initial estimation. The traditional, faster iterative search is used to finetune until convergence.

\section{Experiments}
\label{sec:results}

In this section, we report the results of our work. We first analyze the results of our proposed networks in Section~\ref{sec:netresults}. Then we discuss the results of our resampling method in Section~\ref{sec:resampresults}. In Section~\ref{sec:compresults}, we show qualitative and quantitative comparisons of our approach to previous methods for correcting specific distortion types. In Section~\ref{sec:appresults}, we show some applications of our method. CNNs training details are given in the supplementary material.

\subsection{Networks}
\label{sec:netresults}

\begin{table*}
\begin{center}
\begin{tabular}{llcccccccc}
    \toprule
    \multicolumn{2}{c}{Configuration} & \multicolumn{7}{c}{EPE} \\
    \cmidrule(lr){1-2} \cmidrule(lr){3-9}
    Architecture & Training dataset & $\mathcal{B}$ & $\mathcal{P}i$ & $\mathcal{R}$ & $\mathcal{S}$ & $\mathcal{P}$ & $\mathcal{W}$ & Average \\
    \cmidrule(lr){1-1} \cmidrule(lr){2-2} \cmidrule(lr){3-9} 
    GeoNetS & Single-type  & 1.43 & 0.79 & 2.41 & 2.19 & 0.89 & 1.06 & 1.46 \\
    GeoNetM w/o Clas & Single-type & 1.57 & 1.12 & 3.01 & 2.91 & 1.01 & 1.32 & 1.82 \\
    GeoNetM w/o Clas & Multi-type & 3.07 & 2.24 & 3.75 & 4.99 & 3.35 & 1.73 & 3.19 \\
    GeoNetM & Multi-type & 2.72 & 2.03 & 3.68 & 3.12 & 3.29 & 1.67 & 2.75 \\
    GeoNetM w/ Hou & Multi-type & 1.78 & 1.34 & 2.77 & 2.27 & 2.25 & 1.22 & 1.94 \\
    \bottomrule
\end{tabular}
\end{center}
\caption{EPE and classification statistics of our approach using 500 test images per distortion.}
\label{tab:models}
\end{table*}

\begin{figure*}
    \centering
    \footnotesize
    \includegraphics[width=\linewidth]{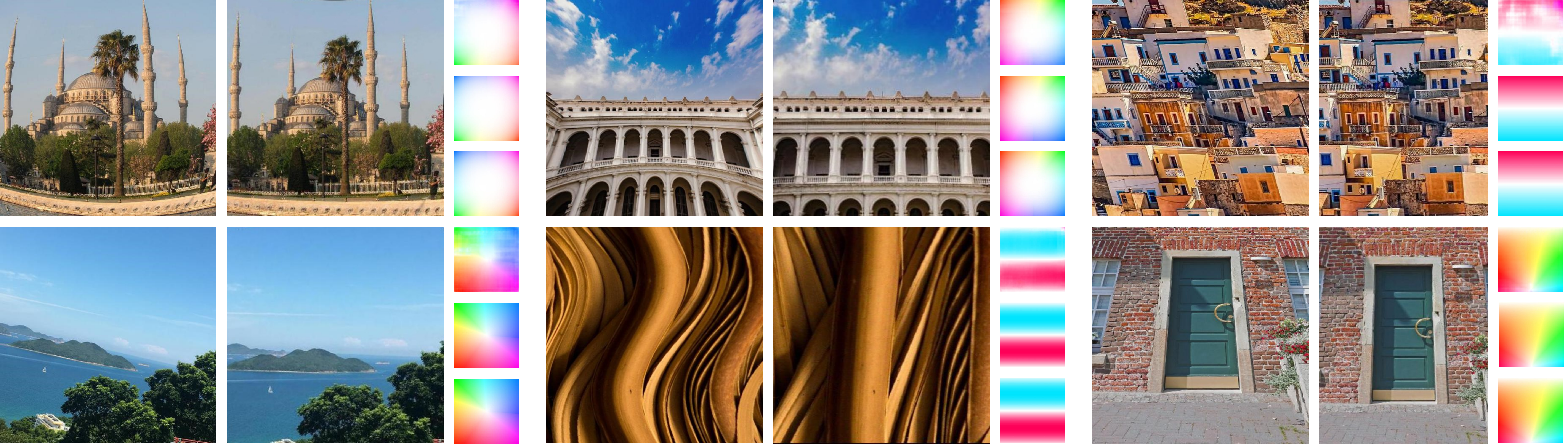}
    \begin{tabular}{x{0.105\linewidth}x{0.105\linewidth}x{0.05\linewidth}x{0.11\linewidth}x{0.11\linewidth}x{0.05\linewidth}x{0.12\linewidth}x{0.08\linewidth}x{0.04\linewidth}}
        Source & Corrected & Flow & Source & Corrected & Flow & Source & Corrected & Flow\\
    \end{tabular}
    \caption{Results of distortions that we considered. Top row: Barrel distortion, pincushion distortion and shear distortion. Bottom row: rotation, wave distortion and perspective distortion. The flows refer to the flow before model fitting (top), after model fitting (middle), and ground truth (bottom).}
    \label{fig:results}
\end{figure*}

To evaluate the performance of GeoNetS, we compare with GeoNetM without classification branch and train these two networks on the same dataset with only single-type distortion. The first two rows in Table~\ref{tab:models} show that by explicitly considering the distortion model in the network, GeoNetS achieves better results. Moreover, the flow is globally smooth due to the restriction given by the distortion model. 

Second, we examine how the joint learning with classification improves the distortion flow prediction. The third and fourth rows in Table~\ref{tab:models} show that GeoNetM with joint learning using the classification branch has more accurate prediction than GeoNetM w/o classification training in the multi-type distortion dataset. The classification accuracy of GeoNetM is $97.3\%$ for these six kinds of distortion. Table~\ref{tab:models} also shows that single-type achieves lower flow error than multi-type since additional information needs to be learned for the multi-type task.  

We also examine whether the model fitting method improves prediction accuracy for GeoNetM. The last two rows in Table~\ref{tab:models} shows that the Hough transform based model fitting provides more accurate results. More results from GeoNetM are shown in Figure~\ref{fig:results}. For each example, the distorted image is shown on the left, the corrected output image in the middle, and the three flows (before fitting, after fitting, and ground truth) on the right. More real image results and detailed discussion of model fitting, GeoNetS and GeoNetM are given in the supplementary material. 

\subsection{Resampling}
\label{sec:resampresults}

\begin{figure}
    \centering
    \includegraphics[width=\linewidth]{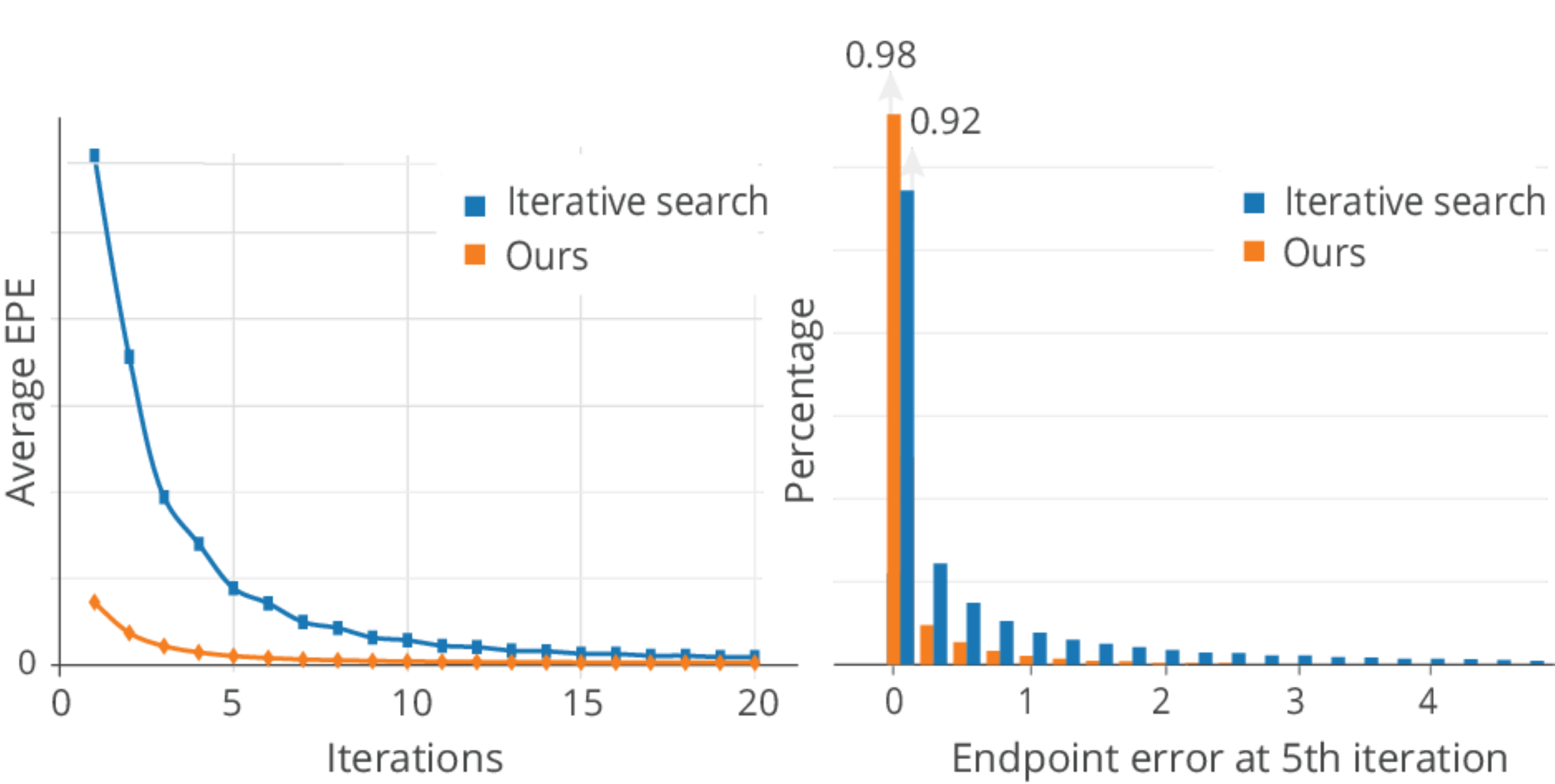}
    \caption{Convergence of EPE (left) and its histogram after five iterations (right) for traditional iterative search and our method. Test on 10 images with different distortion levels.}
    \label{fig:IS_statistics3}
\end{figure}

Next we present the results of our resampling strategy. Figure~\ref{fig:IS_images} shows results applied to images with two different distortion levels. Note that, on the top row, it takes roughly 10 iterations to converge using the traditional iterative search approach (IS), whereas 5 iterations suffice when using our initialization. On the second row, with a more severe distortion, even after 15 iterations, the traditional method has not satisfactorily converged, whereas with our initialization the results also converge within 5 iterations. 

Figure~\ref{fig:IS_statistics3} demonstrates how our method more quickly converges to the ground truth (left), and how the vast majority of pixels already have an endpoint error lower than 1/5 of a pixel after just 5 iterations. A parallel version of our approach has been implemented on the GPU using an Intel Xeon E5-2670 v3 2.3 GHz machine with Nvidia Tesla K80 GPU. It can resample the image under 50 ms. 

\subsection{Comparison with previous techniques}
\label{sec:compresults}

Next, we compare our distortion correction technique to some existing methods that are specialized to some distortion types. Note that, unlike these methods, our learning-based approach is able to handle different distortion types.

\begin{figure}
    \centering
    \footnotesize
    \includegraphics[width=\linewidth]{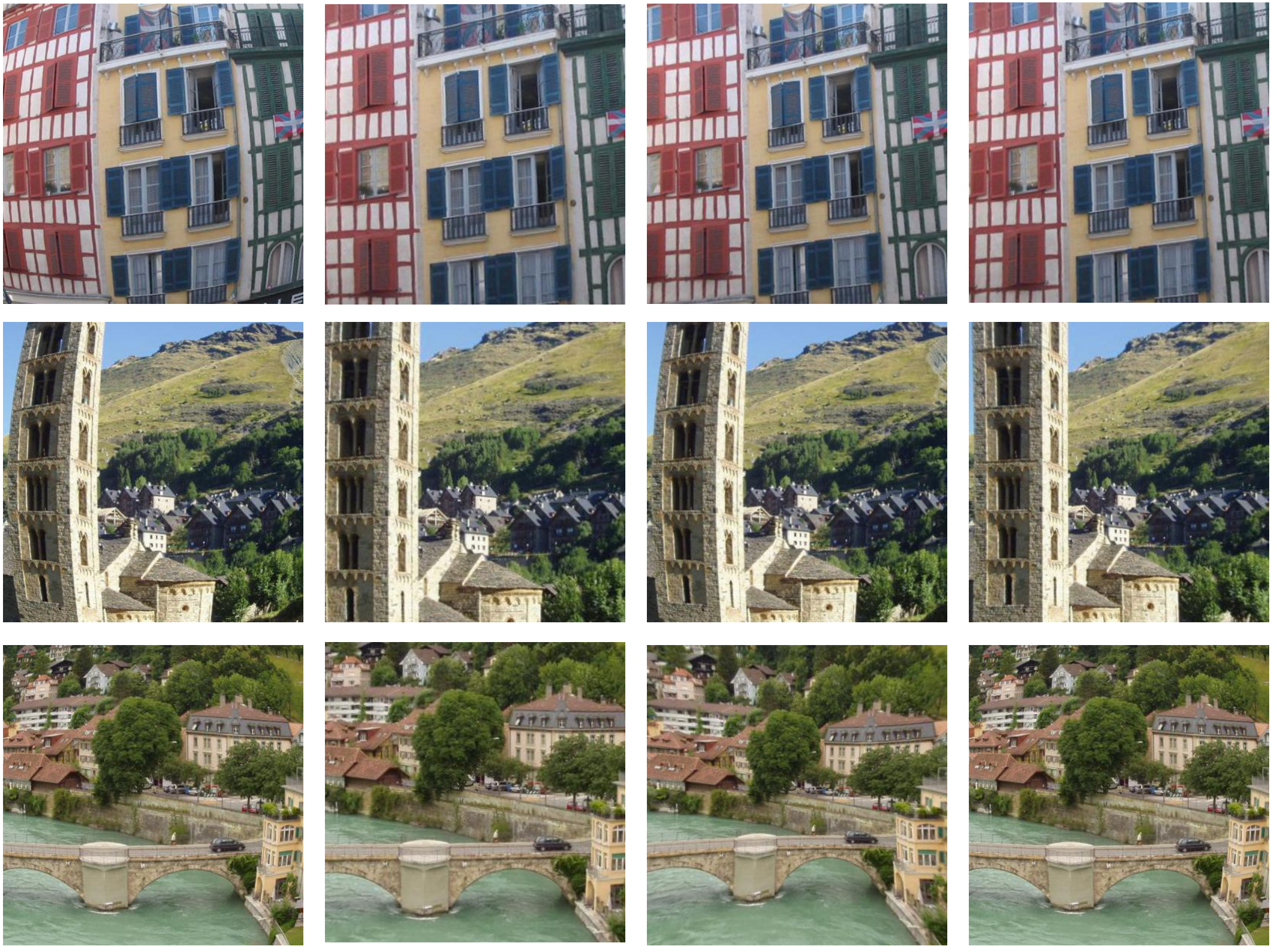}
    \begin{tabular}{x{0.18\linewidth}x{0.21\linewidth}x{0.21\linewidth}x{0.20\linewidth}}
        Source & Ours & Result of~\cite{aleman2014automatic} & Result of~\cite{santana2015invertibility}\\
    \end{tabular}
    \caption{Qualitative comparison with state-of-the-art lens distortion correction methods.}
    \label{fig:compare_imgs}
\end{figure}

\begin{figure}
    \centering
    \includegraphics[width=.8\linewidth]{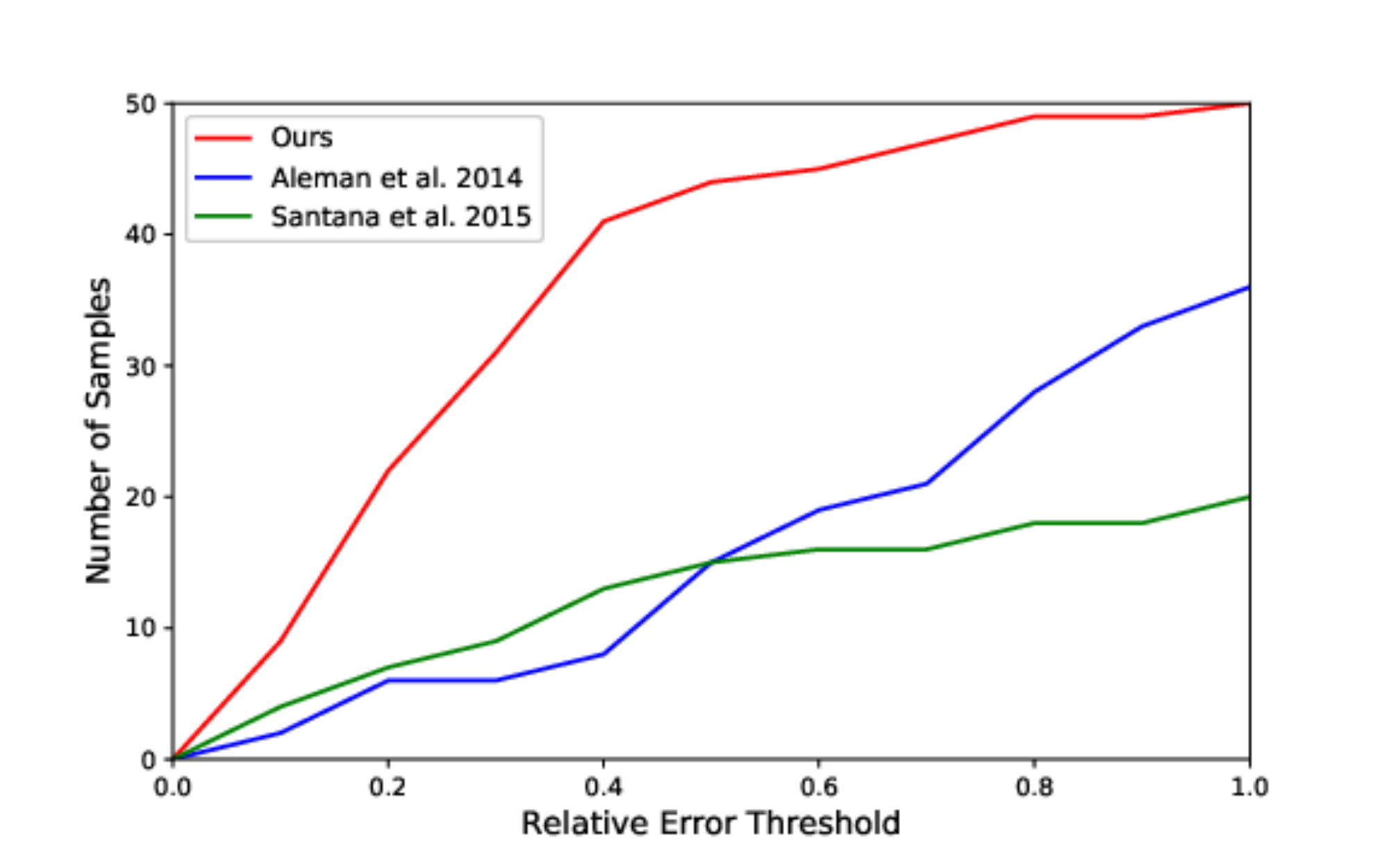}
    \caption{Quantitative comparison on lens distortion correction methods. Given relative error threshold, our method gives more accurate pixels than the other methods.} 
    \label{fig:Barrel_compararison}
\end{figure}

\paragraph{Lens distortion.}
Figure~\ref{fig:compare_imgs} compares our approach with~\cite{aleman2014automatic} and~\cite{santana2015invertibility}, which are specialized for lens distortion. Note that for cases where the image has obvious distortion (e.g., first row), all methods can correct accurately. However, in cases where the distortion is more subtle or does not exhibit highly distorted lines (e.g., bottom two rows), our approach yields improved results. Figure~\ref{fig:Barrel_compararison} shows a quantitative comparison based on 50 randomly chosen images from the dataset of~\cite{jegou2008hamming}. These images include a variety of scene types (e.g., nature, man-made, water) and are distorted with random distortion parameters to generate our synthetic dataset. All of these methods use Fitzgibbon's single parameter division model~\cite{fitzgibbon2001simultaneous}, therefore we can calculate the relative error of the distortion correction parameters for comparison. Note that, with our approach, the number of sample images (y-axis) that lie within the error thresholds (x-axis) is significantly higher than the other methods.



\begin{table}
\centering
\caption{Angle deviation of detected lines with the vertical angle.}
\label{tab:comp_pers}
\begin{tabular}{lccc}
    \toprule
    method &  baseline (input)  &     ~\cite{chaudhury2014auto} & ours \\
    \cmidrule(lr){1-4}
    angle deviation  & $6.44^{\circ}$  & $3.03^{\circ}$  & $2.81^{\circ}$ \\
    \bottomrule
\end{tabular}
\end{table}

\begin{figure}
    \centering
    \small
    \includegraphics[width=\linewidth]{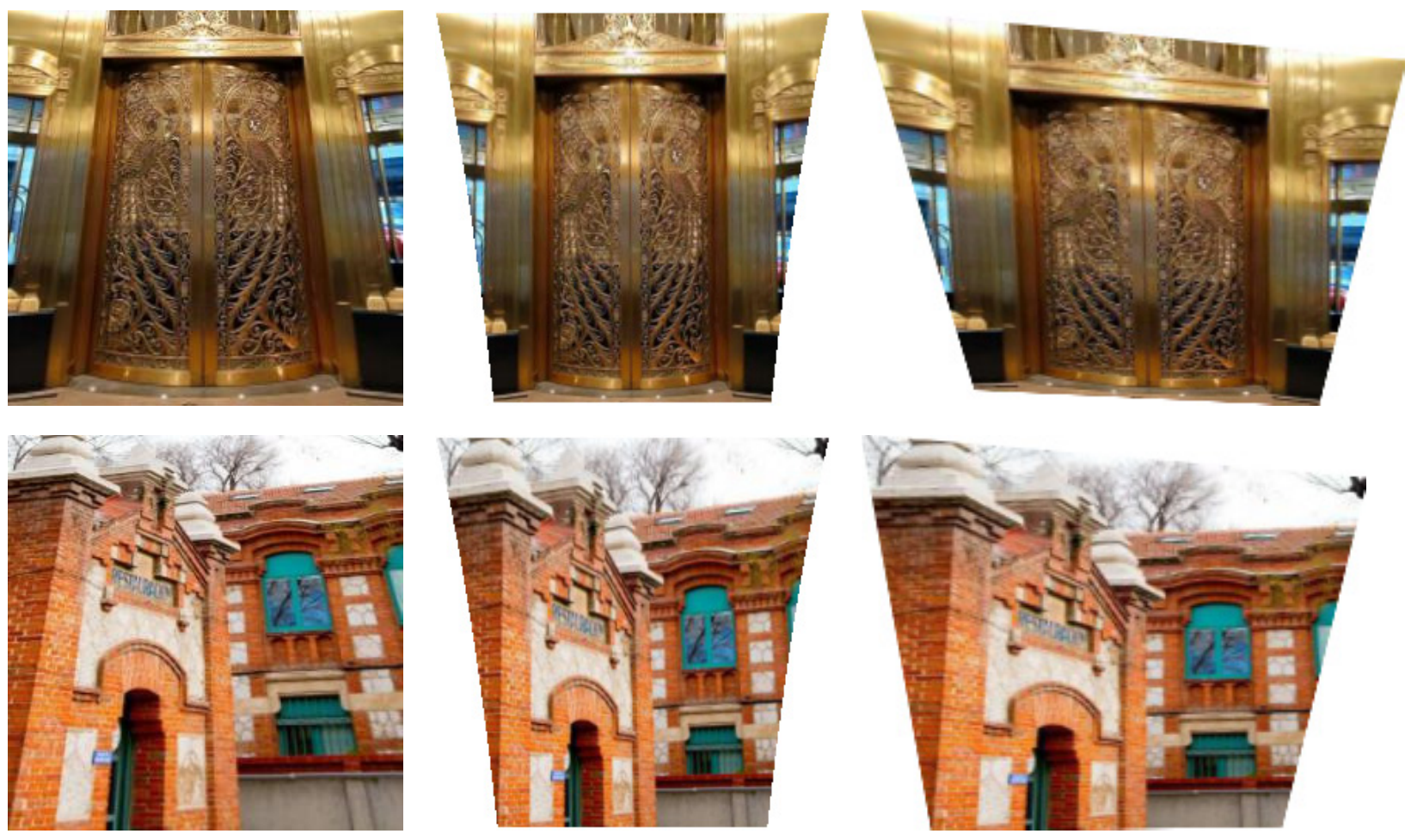}
    \begin{tabular}{x{0.2\linewidth}x{0.3\linewidth}x{0.3\linewidth}}
        Source & Ours & Result of~\cite{chaudhury2014auto} \\
    \end{tabular}
    \caption{Comparison with previous perspective correction method.}
    \label{fig:compare_per}
\end{figure}

\paragraph{Perspective distortion.}
For the perspective distortion, we compare with~\cite{chaudhury2014auto}. Here we use angle deviation as the metric. We collect 30 building images under orthographic projection and distort them with different Homography matrices. We control the distorted vertical lines within $[70^{\circ}, 110^{\circ}]$. Then we detect straight lines~\cite{von2010lsd} in the correction results using the line segment detector within this range and assume that the angle of these lines should be $90^{\circ}$ after correction and calculate their average angle deviation. As shown in Table~\ref{tab:comp_pers} and Figure~\ref{fig:compare_per}, our method outperforms the previous approach~\cite{chaudhury2014auto}.

\subsection{Applications}
\label{sec:appresults}

In addition, we explored applications that can benefit from our distortion correction method directly.

\begin{figure}
    \centering
    \small
    \includegraphics[width=\linewidth]{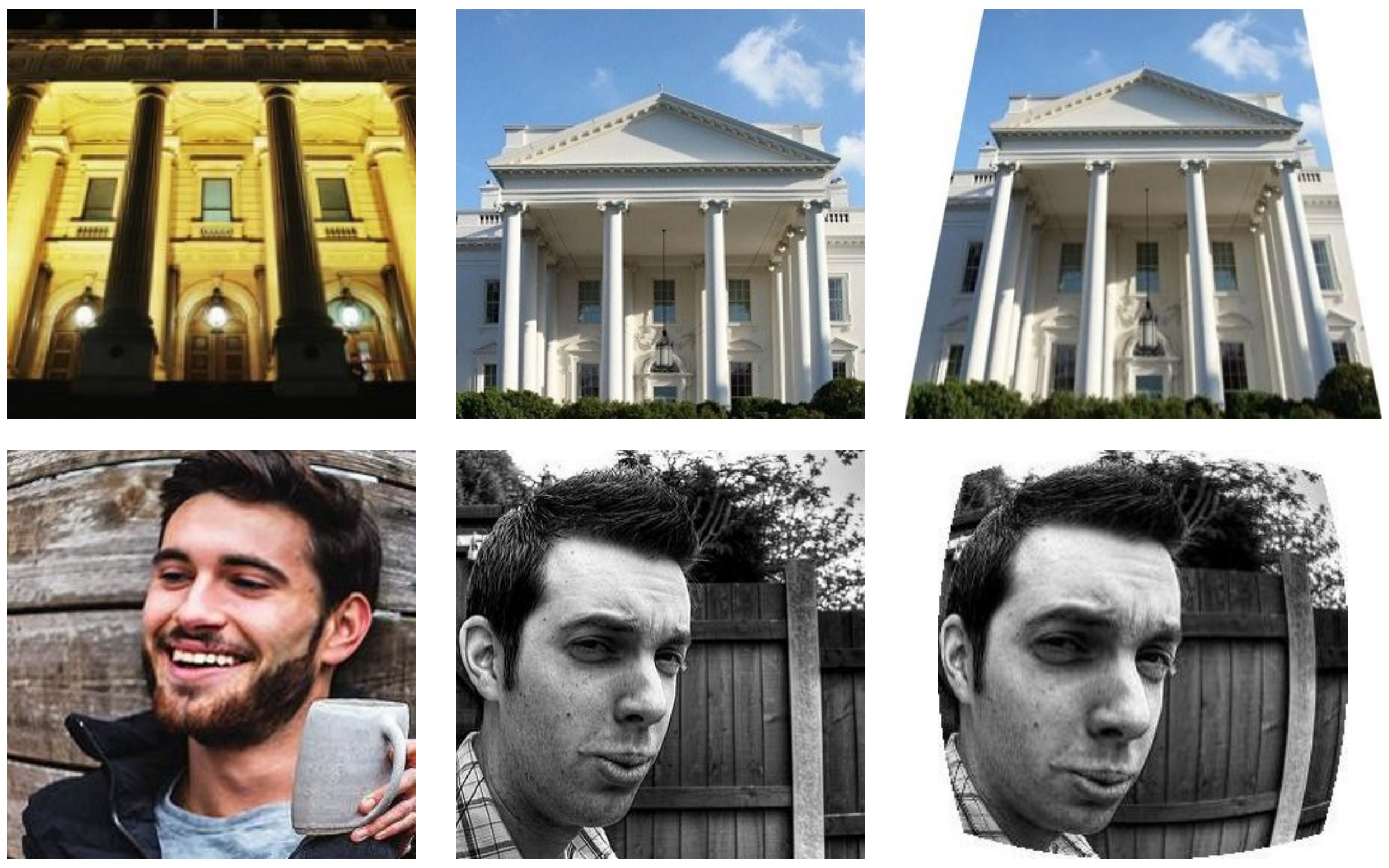}
        \begin{tabular}{x{0.25\linewidth}x{0.29\linewidth}x{0.3\linewidth}}
        Reference & Target & Result \\
    \end{tabular}
    \caption{Examples of distortion transfer. Perspective and barrel distortions are used.}
    \label{fig:transfer}
\end{figure}

\paragraph{Distortion transfer.} Our system can detect the distortion from a reference image and transfer to a target image. We can estimate the forward flow from the reference image to the corrected version and then directly apply it to the target by bilinear interpolation. Figure~\ref{fig:transfer} shows two examples of transferring distortion from a reference image to a target image, in order to accentuate the perspective of a house photograph (upper row) or apply aggressive barrel distortion to a portrait (lower row).

\begin{figure}
    \centering
    \footnotesize
    \includegraphics[width=\linewidth]{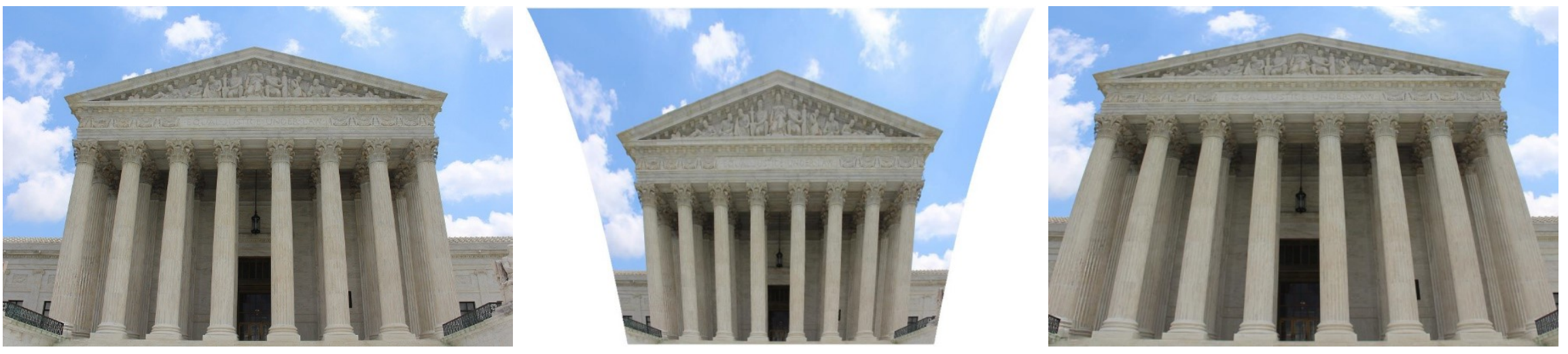}
    \begin{tabular}{x{0.27\linewidth}x{0.31\linewidth}x{0.3\linewidth}}
        Source & Shrunk & Expanded\\
    \end{tabular}
    \caption{Example of distortion exaggeration.}
    \label{fig:exaggerate}
\end{figure}
\paragraph{Distortion exaggeration.} To achieve distortion exaggeration, we can reverse the direction of estimated flow field to make the pixels further away from its undistorted position, and use our resampling approach to generate an exaggerated distortion output. Figure~\ref{fig:exaggerate} shows a building with perspective effect, we can adjust the level of distortion to exaggerate the effect by amplifying or reversing the flow, respectively.

\paragraph{Co-occurring distortion correction.}
Sometimes an image could have more than one type of distortion. We can correct the distorted image simply by running our correction algorithm twice iteratively. For each iteration, it detects and corrects the most severe type of distortion that it encounters. See the supplementary material for some examples and results.

\section{Conclusion}
In conclusion, we present the first approach to blindly correct several types of geometric distortions from a single image. Our approach uses a deep learning method trained on several common distortions to detect the distortion flow and type. Our model fitting and parameter estimation approach then accurately predicts the distortion parameters. Finally, we present a fast parallel approach to resample the distortion-corrected images. We compare our techniques to recent specialized methods for distortion correction and present applications such as distortion transfer, distortion exaggeration, and co-occurring distortion correction.

{
\small
\bibliographystyle{ieee}
\bibliography{egbib}
}

\end{document}